\documentclass[runningheads]{llncs}
\usepackage[dvips,final]{graphicx}
\usepackage[top=2.5cm, bottom=2cm, left=2.7cm, right=2.7cm]{geometry}

\usepackage{makecell}
\usepackage{enumerate}
\usepackage{color}
\usepackage{caption}
\usepackage{subcaption}
\usepackage{amsfonts}
\usepackage{amsmath}
\usepackage{amssymb}
\usepackage{times}
\usepackage{cite}
\usepackage{hhline}
\usepackage{compactbib}
\usepackage{graphicx}        
\usepackage{amsmath,amsfonts,amssymb}
\usepackage{geometry}
\usepackage{lineno,hyperref}
\usepackage{mathrsfs}
\usepackage[labelsep=period]{caption}
\usepackage{tikz}
\usetikzlibrary{shapes,arrows}
\usetikzlibrary{bayesnet}
\usetikzlibrary{fit,positioning}
\usepackage{pdfpages}
\usepackage{bm}

\definecolor{halfgreen}{RGB}{0,128,0}
\definecolor{ahsred}{RGB}{192,0,0}

\renewcommand{\thefigure}{\arabic{figure}}
\renewcommand{\thetable}{\Roman{table}}

\newcommand{\beq}{\begin{equation}}
	\newcommand{\eeq}{\end{equation}}
\newcommand{\bgqar}{\begin{eqnarray}}
	\newcommand{\enqar}{\end{eqnarray}}
\newcommand{\bgqarn}{\begin{eqnarray*}}
	\newcommand{\enqarn}{\end{eqnarray*}}
\newcommand{\bgary}{\begin{array}}
	\newcommand{\enary}{\end{array}}

\renewcommand{\refname}{}
\begin{document}
\title{Quantifying the value of positive transfer: An experimental case study}
%
%
\author{Aidan J.\ Hughes\inst{1} \and
Giulia Delo\inst{2} \and
Jack Poole\inst{1} \and
Nikolaos Dervilis\inst{1} \and
Keith Worden\inst{1}}
\authorrunning{A.J.\ Hughes et al.}
%
\institute{Dynamics Research Group, Department of Mechanical Engineering, University of Sheffield, Sheffield, S1 3JD, UK \and
Department of Mechanical and Aerospace Engineering, Politecnico di Torino, 10129 Turin, Italy
\email{aidan.j.hughes@sheffield.ac.uk}}
\maketitle              
\begin{abstract}
In traditional approaches to structural health monitoring, challenges often arise associated with the availability of labelled data. Population-based structural health monitoring seeks to overcomes these challenges by leveraging data/information from similar structures via technologies such as transfer learning. The current paper demonstrate a methodology for quantifying the value of information transfer in the context of operation and maintenance decision-making. This demonstration, based on a population of laboratory-scale aircraft models, highlights the steps required to evaluate the expected value of information transfer including similarity assessment and prediction of transfer efficacy. Once evaluated for a given population, the value of information transfer can be used to optimise transfer-learning strategies for newly-acquired target domains.

\end{abstract}
\section{Introduction}

Population-based SHM (PBSHM) \cite{Bull2021,Gosliga2021,Gardner2021b,Tsialiamanis2021} seeks to overcome one of the fundamental challenges with traditional data-based SHM – a scarcity of comprehensively-labelled datasets. PBSHM leverages information-sharing technologies such as transfer learning or multi-task learning to improve predictive capabilities in data-scarce target domains using information from data-rich source domains. Transfer-learning techniques, such as domain adaptation, have been demonstrated to be effective at enabling information-sharing within structural populations, e.g. bridges \cite{Gardner2022domain}. However, transfer learning between given source and target domains is not guaranteed to be successful; in some cases, performing transfer learning will degrade predictive capabilities in the target domain – this phenomenon is known as \textit{negative transfer} \cite{Wang2019negative}.

There is strong motivation to maximise the effectiveness of transfer learning in PBSHM when one considers that SHM systems are primarily decision-support tools, informing operation and maintenance (O\&M) strategies for high-value and safety-critical assets. When framed in the context of O\&M decision-making, misclassifications for the presence, location, or extent of damage correspond to undesirable outcomes such as unnecessary inspections, or even missed repairs and structural failures. Needless to say, these outcomes incur some cost to the operator and would ideally be avoided.

To this end, a decision-theoretic approach for optimising transfer-learning strategies has been developed. This approach is based on the expected value of information transfer (EVIT) \cite{Hughes2024evit}. The current paper aims to demonstrate how the EVIT can be assessed between a target domain and a population of source domains comprising experimental structures. To achieve this aim, post-transfer prediction quality for the given target domain will be forecast by considering a measure of structural similarity with candidate source domains. This measure of similarity will be assessed by considering characteristics of the source and target domain structures such as topology, density, and scale.

\section{Background Theory}

\subsection{Transfer Learning for PBSHM}
Transfer-learning techniques aim use information from a data-, or label-, rich \textit{source domain} $\mathcal{D}_s = \{ \mathbf{x}_{s,i}, y_{s,i} \}_{i=1}^{N}$, in order to improve predictions in a data-, or label-, scarce \textit{target domain} $\mathcal{D}_t = \{\mathbf{x}_{t,j},y_{t,j}\}_{j = 1}^{M}$. In the context of PBSHM, it is assumed that there is a subset of individual structures within a population for which labelled data are available --- these structures can be considered as candidate source domains. It then follows that the individuals within a population for which labelled data are unavailable can be considered as target domains \cite{Gardner2021b}. When applying transfer learning, it is typically assumed that the marginal distributions of observable data differ, i.e.\ $P(X_s) \neq P(X_t)$, and/or that the conditional distributions of labels differ, i.e.\ $P(\mathbf{y}_s|X_s) \neq P(\mathbf{y}_t|X_t)$. In SHM applications, these discrepancies between source and target domain structures arise because of factors such as manufacturing variability; geometric, topological, and material differences; and operational and environmental differences \cite{Gardner2021b}.

Several transfer-learning approaches have been applied to PBSHM. Various flavours of \textit{domain adaptation} have been used to harmonise source and target domains in the context of PBSHM  \cite{Gardner2018adaptation,Poole2022alignment}. Other popular transfer-learning techniques for PBSHM include neural approaches \cite{Soleimani2023zeroshot,Cao2018preprocessing} and multi-level modelling \cite{Dardeno2023hierarchical}.

An important consideration when applying transfer-learning techniques in PBSHM is the possibility of negative transfer \cite{Wang2019negative}. Negative transfer is characterised by a degradation in predictive performance in the target domain, following an information transfer attempt. Negative transfer is known to occur if the source joint distribution $P(X_s,\mathbf{y}_s)$ and target joint distribution $P(X_t, \mathbf{y}_t)$ are not sufficiently similar; or, if the algorithm used to conduct transfer is unable to find the correct mapping for other reasons, such as non-uniqueness of solutions.

\subsection{Value of Information Transfer}

Value of information (VoI) is a concept in decision theory defined to be the amount of resource (often money) a decision-maker should be willing to expend in order to gain access to information prior to making a decision. VoI has been applied to traditional SHM in recent works \cite{Kamariotis2022,Hughes2022,Hughes2022c,Zhang2023quantifying}. In \cite{Hughes2023decision}, value of information was leveraged in the context of PBSHM; specifically, to quantify the benefits of applying transfer learning techniques in a decision-theoretic manner. Termed the \textit{expected value of information transfer} (EVIT), this quantity can be interpreted as the price a decision-maker should be willing to pay in order to transfer information from a source domain prior to making decisions in a given target domain.

As detailed in \cite{Hughes2023decision}, the EVIT can be computed as follows,

\begin{equation}
	\text{EVIT}(\mathcal{T}) = \text{EU}(\mathcal{Q}|\mathcal{T}) - \text{EU}(\mathcal{Q}|\mathcal{T}_0) = P(\mathcal{Q}|\mathcal{T})\cdot U(\mathcal{Q}) - P(\mathcal{Q}|\mathcal{T}_0)\cdot U(\mathcal{Q}),
	\label{eq:EVIT}
\end{equation}
\noindent where $\text{EU}$ denotes the expected utility. $\mathcal{T}$ denotes a given transfer strategy parametrised by a source domain $\mathcal{D}_s$ and a transfer algorithm $\mathcal{A}$, with $\mathcal{T}_0$ corresponding to a null transfer strategy representing the case where no transfer is performed (i.e.\ $\mathcal{D}_s = \emptyset$ and $\mathcal{A} = \mathbb{I}$, where $ \mathbb{I}$ is an identity function). $\mathcal{Q}$ denotes some set of prediction quality criteria that can be related to utilities/costs in the context of a target-domain decision process via the utility function $U(\mathcal{Q})$. 

Here, it is worth noting that equation (\ref{eq:EVIT}) implies that $\text{EVIT}(\mathcal{T}_0) = 0$. It follows that one can define negative and positive transfer from a decision-theoretic viewpoint -- simply that negative transfer is anticipated when $\text{EVIT}(\mathcal{T}) < 0$, and positive transfer is anticipated when $\text{EVIT}(\mathcal{T}) > 0$. Considering equation (\ref{eq:EVIT}) and the associated definition of negative transfer, one can maximise the expected benefits of information transfer via the optimisation,

\begin{equation}
	\mathcal{T}^{\ast} = \mathcal{T}(\mathcal{D}_{s}^{\ast}, \mathcal{A}^{\ast}) = \underset{\mathcal{T}}{\text{argmax}} \big[ \text{EVIT}(\mathcal{T}) + U(\mathcal{T}) \big],
	\label{eq:T_opt}
\end{equation}

\noindent where an asterisk superscript denotes an optimal strategy/parameter, and $U(\mathcal{T})$ is a utility function specifying the cost associated with executing a given transfer strategy. Here, $\mathcal{T}^{\ast}$ comprises the source domain and transfer algorithm that minimise the risk of negative transfer.

In practice, to conduct the optimisation stated in equation (\ref{eq:T_opt}), one must forecast the post-transfer prediction qualities for target-domain data. To make such predictions, one can leverage a key assumption underpinning PBSHM; specifically, that members of a population that are structurally similar should yield improved transfer results in comparison to members of the population that are structurally disparate. By considering pair-wise transfers between candidate source domains, one can construct a data set consisting of features based on structural similarity, and transfer efficacy, from which a mapping can be learned \cite{Poole2023negative}. Thus far, the approach for computing the EVIT has been demonstrated using only a numerical case study. The following section introduces an experimental case study to demonstrate the approach.

\section{Case Study: GARTEUR Population}

To demonstrate the approach for quantifying the EVIT, an experimental case study is presented in the current section. The population used for this case study comprises eight laboratory-scale aircraft models. These structures are based on the Group for Aeronautical Research Technology in Europe (GARTEUR) benchmarking testbed detailed in \cite{Balmes1997}. Variability is present within the population; arising from both topological differences and differences in attributes such as construction material and overall scale. A summary of the population is provided in Table \ref{tab:garteurs}. There are three topologies of structure within the population. The first type, referred to as `base' corresponds to a simplified aircraft structure comprising a fuselage, tailplane and wing. The second type, denoted `+winglets', is almost identical to base type however possess additional winglets at the tips of the wings. The third type, denoted `+engines', is again almost identical to the base type, however possesses masses part-way along the wings, simulating engines. The population comprises structures of two scales; the structures denoted as `large' have wingspans of 2m, whereas the structures denoted `small' have wingspans of 1m with other dimensions scaled proportionally. For a more comprehensive description of the population, the reader is directed to \cite{Delo2023influence}.

\begin{table}[h!]
	\centering
	\caption{Summary of the GARTEUR population. Young's modulus is denoted by $E$ and mass density is denoted by $\rho$.}
	\begin{tabular}{lllllll}
		\cline{1-6}
		\multicolumn{1}{|l|}{\textbf{ID}} & \multicolumn{1}{c|}{\textbf{Topology}} & \multicolumn{1}{c|}{\textbf{Scale}} & \multicolumn{1}{c|}{\textbf{Material}} & \multicolumn{1}{c|}{$E$ (GPa)} & \multicolumn{1}{c|}{$\rho$ (kgm$^{-3}$)} & \\ \cline{1-6}
		\multicolumn{1}{|l|}{G1} & \multicolumn{1}{c|}{+winglets} & \multicolumn{1}{c|}{small} & \multicolumn{1}{c|}{brass} & \multicolumn{1}{c|}{90} & \multicolumn{1}{c|}{8400} & \\ \cline{1-6}
		\multicolumn{1}{|l|}{G2} & \multicolumn{1}{c|}{+winglets} & \multicolumn{1}{c|}{large} & \multicolumn{1}{c|}{aluminium} & \multicolumn{1}{c|}{68} & \multicolumn{1}{c|}{2710} & \\ \cline{1-6}
		\multicolumn{1}{|l|}{G3} & \multicolumn{1}{c|}{+winglets} & \multicolumn{1}{c|}{large} & \multicolumn{1}{c|}{steel} & \multicolumn{1}{c|}{200} & \multicolumn{1}{c|}{8000} & \\ \cline{1-6}
		\multicolumn{1}{|l|}{G4} & \multicolumn{1}{c|}{+winglets} & \multicolumn{1}{c|}{large} & \multicolumn{1}{c|}{aluminium} & \multicolumn{1}{c|}{68} & \multicolumn{1}{c|}{2710} & \\ \cline{1-6}
		\multicolumn{1}{|l|}{G5} & \multicolumn{1}{c|}{+engines} & \multicolumn{1}{c|}{small} & \multicolumn{1}{c|}{aluminium} & \multicolumn{1}{c|}{68} & \multicolumn{1}{c|}{2710} & \\ \cline{1-6}
		\multicolumn{1}{|l|}{G6} & \multicolumn{1}{c|}{base} & \multicolumn{1}{c|}{small} & \multicolumn{1}{c|}{steel+composite} & \multicolumn{1}{c|}{250} & \multicolumn{1}{c|}{3000} & \\ \cline{1-6}
		\multicolumn{1}{|l|}{G7} & \multicolumn{1}{c|}{base} & \multicolumn{1}{c|}{small} & \multicolumn{1}{c|}{steel} & \multicolumn{1}{c|}{200} & \multicolumn{1}{c|}{8000} & \\ \cline{1-6}
		\multicolumn{1}{|l|}{G8} & \multicolumn{1}{c|}{base} & \multicolumn{1}{c|}{large} & \multicolumn{1}{c|}{steel} & \multicolumn{1}{c|}{200} & \multicolumn{1}{c|}{8000} & \\ \cline{1-6}
	\end{tabular}
	\label{tab:garteurs}
\end{table}

Data were acquired from the structures summarised in Table \ref{tab:garteurs} via modal tests considering the undamaged structures and several simulated damage scenarios. The damage scenarios were simulated by adding masses at several locations across the structure; specifically, three locations along the length of one wing, one location on the tailplane, and two locations on the fuselage. From these modal tests, labelled datasets for each individual structure were constructed by corrupting the mean natural frequencies extracted with Gaussian noise.

To construct a PBSHM case study around the GARTEUR population datasets, the classification tasks of interest, the transfer tasks of interest, and the methodology for assessing structural similarity must be defined.

\subsection{Classification Task}

The classification task used in the current case study is framed as a SHM damage detection and localisation problem. Specifically, the task is to use the first two natural frequencies $\bm{x} \in \mathbb{R}^{2}_{+} $ to predict one of four structural health states $y \in \{0,1,2,3\}$, where $y=0$ corresponds to the undamaged health state, $y=1$ corresponds to wing damage, $y=2$ corresponds to tailplane damage, and $y=3$ corresponds to fuselage damage. To establish a mapping $f: X \rightarrow Y$, a $k$-nearest neighbours classifier was used with $k=5$.

To evaluate predictive performance, four measures were considered; the true prediction rate (TR), the false positive prediction rate (FPR), the false negative rate (FNR), and the false damage prediction rate (FDR). The TR corresponds to the proportion of correct predictions within a test dataset, the FPR corresponds to the proportion of type-I errors, where a type-I error occurs when damage is predicted when in fact there is none. The FNR corresponds to the proportion of type-II errors, where a type-II error occurs when no damage is predicted when in fact there is some damage present. Finally, FDR corresponds to the proportion of a third error type whereby damage is correctly detected however is localised incorrectly. The prediction quality measures can be summarised as a vector $\bm{q} = \{\text{TR}, \text{FPR}, \text{FNR}, \text{FDR}\}$.

These prediction quality measures were chosen as they are agnostic to the size of the target-domain dataset but also because they can be easily interpreted in the context of SHM decision-making. TR-type predictions would correspond to optimal O\&M actions being selected as damage, or absence of damage, would be correctly identified; thus, predictions of this type would correspond to positive utility. FPR-type predictions would result in unnecessary O\&M actions being undertaken; while this doesn't undermine the integrity of the structure, resources would be wasted and therefore predictions of this type would correspond to negative utility. FNR-type predictions would result in inaction for cases where intervention is necessary meaning the integrity of the system is undermined potentially resulting in catastrophic failure; consequently, this type of prediction would correspond to large negative utility in an SHM decision process. Finally, FDR-type predictions yield correct detection of damage therefore the resulting O\&M actions would not necessarily adversely affect the structural integrity, however information pertaining to the location of such damage would be misleading and would likely result in additional resources being expended to track down the source of damage; as such, this prediction type warrants small negative utility. For the purposes of the current case study, the mapping from prediction quality to utility in the context of O\&M is specified by the utility function presented in Table \ref{tab:UQ}.

\begin{table}[h!]
	\centering
	\caption{The utilities associated with the prediction types True, False Positive, False Negative, and False Damage.}
	\begin{tabular}{llllll}
		\cline{1-5}
		\multicolumn{1}{|l|}{\textbf{Prediction Type}} & \multicolumn{1}{c|}{True} & \multicolumn{1}{c|}{False Positive} & \multicolumn{1}{c|}{False Negative} & \multicolumn{1}{c|}{False Damage} & \\ \cline{1-5}
		\multicolumn{1}{|l|}{\textbf{Utility}} & \multicolumn{1}{c|}{5} & \multicolumn{1}{c|}{-10} & \multicolumn{1}{c|}{-50} & \multicolumn{1}{c|}{-5} & \\ \cline{1-5}
	\end{tabular}
	\label{tab:UQ}
\end{table}

\subsection{Transfer Task}

For the GARTEUR population, there are 56 unique transfer tasks that can be considered as each structure can be used as a source domain for every structure other than itself. Each transfer task can be constructed taking a pair of structures and obscuring the damage labels for one of the structures such that it can be considered a target domain. 

For the current case study, a simple, but often highly-effective form of domain adaptation known as normal-condition alignment (NCA) is considered \cite{Poole2022alignment}. In NCA, data are translated and scaled such that the undamaged data in the target domain aligns with the undamaged data in the source domain. Assuming source domain data are standardised, NCA is given by,

\begin{equation}
	z_t^{(j)} = \bigg(\frac{x_t^{(j)} - \mu_{t,n}}{\sigma_{t,n}}\bigg) \sigma_{s,n} + \mu_{s,n}
\end{equation}

\noindent where $z_t$ are the transformed target-domain features, $x_t$ are the original target-domain features, and $\mu_{t,n}$, $\sigma_{t,n}$, $\mu_{s,n}$, and $\sigma_{s,n}$ are the mean and standard deviation of the normal-condition data for the target and source domains, respectively. Following the application of NCA, predictions can be made for data in the target domain as the kNN classifier can leverage labelled data from the source domain.

Because the labels for each pseudo-target domains were only obscured for the purposes of generating transfer tasks and are in fact available, the post-transfer prediction quality for each task can be assessed by comparing the predicted values with the ground-truth targets. This process generated the data $Q = \{ \mathbf{q}_n \}_{n=1}^{N_T}$ to be used as the targets for establishing the mapping between structural similarity and prediction quality.

\subsection{Similarity Assessment}

As shown in \cite{Poole2023negative}, structural similarity can be used as a predictor of post-transfer prediction quality. For the current case study, a custom similarity measure was constructed based on the $\ell^2$-norm on a Euclidean space $\mathbb{R}^4$, where a vector $\bm{s}_i$ in this four-dimensional space represents a GARTEUR structure $i$ in terms of its topology, scale, Young's modulus, and density. Pairwise distances between structures were evaluated as,

\begin{equation}\label{eq:distance}
	d_{i,j} = ||\bm{s}_i - \bm{s}_j|| = \sqrt{\sum_{k}(s_i^k - s_j^k)^2}.
\end{equation}

These distances were normalised to the interval such that $\hat{d}_{i,j} \in [0,1]$. A pairwise similarity $\varsigma_{i,j}$ was then defined as,

\begin{eqnarray}
	\varsigma_{i,j} = 1 - \hat{d}_{i,j}.
\end{eqnarray}

Initially, each of the four characteristics were scaled to the interval $[0,2]$ so that they were equally weighted in the Euclidean distance. The topologies were scaled according to their relative similarity which assessed using the Jaccard index and a graphical representation of each of the structures --- for further details regarding graphical representations of structures and similarity assessment using the Jaccard index, the reader is directed to \cite{Gosliga2021}. It was determined that the most disparate topologies were those of the structures with engines and winglets. Thus, these topologies were mapped to the values of 0 and 2, respectively. The base topology was then assigned a value of 1.

When the transfer efficacy in terms of target-domain prediction quality (TR) was compared with similarity $\varsigma$, it was found that there was little correlation between the quantities. This finding is at odds with the core tenet of PBSHM that more similar structures yield better transfer and would make it challenging to establish a mapping $\varsigma \rightarrow \bm{q}$; thus, a more informed approach was adopted. It is likely that the negligible correlation was a result of not properly accounting for the importance of each characteristic in the distance metric; for example, one could intuit that for the population of GARTEURS that exhibit a very high degree of topological similarity between individuals, that perhaps this characteristic should not factor in so heavily in the evaluation of the distances. To allow for varying degrees of import over the dimensions of the space, a set of weights were introduced $\bm{w} = \{w_1, w_2, w_3, w_4\}$ to equation (\ref{eq:distance}) such that,

\begin{equation}
	d_{i,j} = ||\bm{s}_i - \bm{s}_j||_{\bm{w}}  = \sqrt{\sum_{k}w_k(s_i^k - s_j^k)^2},
\end{equation}

\noindent where $||\cdot||_{\bm{w}}$ denotes the weighted norm. The weights $\bm{w}$ were determined by performing an optimisation such that the Pearson correlation coefficient $r$ between TR and $\varsigma$ was maximised, under the constraint $||\bm{w}||=1$. It was found that the weights corresponding to topology and Young's modulus went to zero indicating that these characteristics are not important for predicting transfer efficacy within the GARTEUR population. The weights for density and scale were non-zero indicating that, for the current case study, these characteristics are important for determining transfer efficacy.

\subsection{Predicting Transfer Efficacy}

As alluded to earlier, a critical step for evaluating the EVIT is to establish a mapping $g: \varsigma \in \mathcal{S} \rightarrow \bm{q} \in \mathcal{Q}$. This mapping $g$ can be learned from the similarities and prediction qualities generated by considering pairwise transfers between structures as described in the previous subsections. To learn $g$, one must perform a probabilistic vector-valued regression. For the current case study, there are constraints on this regression, specifically:

\begin{itemize}
	\item $0\leq \text{TR, FPR, FNR, FDR} \leq 0$.
	\item $\text{TR} + \text{FPR} + \text{FNR} + \text{FDR} = 1$.
\end{itemize}

\noindent These constraints mean that $\bm{q}$ exists in the 3-simplex; for this reason, the Dirichlet distribution is assumed to be appropriate for modelling $\bm{q}$,

\begin{equation}
	\bm{q}_{n} \sim \text{Dir}(\bm{\alpha}_n)
\end{equation}

\noindent where $\bm{\alpha}_n$ are the Dirichlet distribution concentration parameters. By making this assumption, learning the mapping $g$ amounted to learning a function to regress from the structural-similarity measures $\varsigma_n$ to the latent concentration parameters of the Dirichlet distribution $\bm{\alpha}_n$. To perform this vector-valued regression, the authors opted to use an artificial neural network -- specifically, a multi-layer perceptron (MLP). The architecture of the MLP used is provided in Table \ref{tab:mlp}.

\begin{table}[h!]
	\centering
	\caption{The architecture of the multi-layer perceptron.}
	\begin{tabular}{llllll}
		\cline{1-5}
		\multicolumn{1}{|l|}{\textbf{Layer}} & \multicolumn{1}{c|}{Input} & \multicolumn{1}{c|}{Hidden 1} & \multicolumn{1}{c|}{Hidden 2} & \multicolumn{1}{c|}{Output} & \\ \cline{1-5}
		\multicolumn{1}{|l|}{\textbf{Units}} & \multicolumn{1}{c|}{1} & \multicolumn{1}{c|}{8} & \multicolumn{1}{c|}{8} & \multicolumn{1}{c|}{4} &  \\ \cline{1-5}
		\multicolumn{1}{|l|}{\textbf{Activation}} & \multicolumn{1}{c|}{\texttt{softplus}} & \multicolumn{1}{c|}{\texttt{softplus}} & \multicolumn{1}{c|}{\texttt{softplus}} & \multicolumn{1}{c|}{\texttt{softplus}} &  \\ \cline{1-5}
	\end{tabular}
	\label{tab:mlp}
\end{table}

The loss function used to optimise the weights and biases of the MLP comprised three terms: the first term corresponds to the negative log marginal likelihood of the Dirichlet distribution; the second term imposed a soft preference for monotonic functions to reflect the belief that more similar structures yield better transfer outcomes; and the third term, a function of $||\bm{\alpha}_n||$, penalised overconfident predictions away from data-scarce regions of the $\varsigma$ domain. Further details of this loss function are provided in \cite{Hughes2024evit}.

The Dirichlet distribution predictions obtained using the MLP can be used to estimate the number of each prediction type expected for a target-domain classification task following transfer from a source domain with similarity score $\varsigma$. These predictions can then be combined with the utility function shown in Table \ref{tab:UQ} and equation (\ref{eq:EVIT}) to yield the EVIT from a source domain with similarity $\varsigma$.

To summarise, the methodology detailed in the current section was applied to an experimental case study formulated around a population consisting of eight laboratory-scale aircraft models, or GARTEUR structures. In the next section, the results of this case study will be presented.

\section{Results}

\begin{figure}[h!]
	\centering
	\begin{subfigure}[b]{.45\linewidth}
		\includegraphics[width=\linewidth]{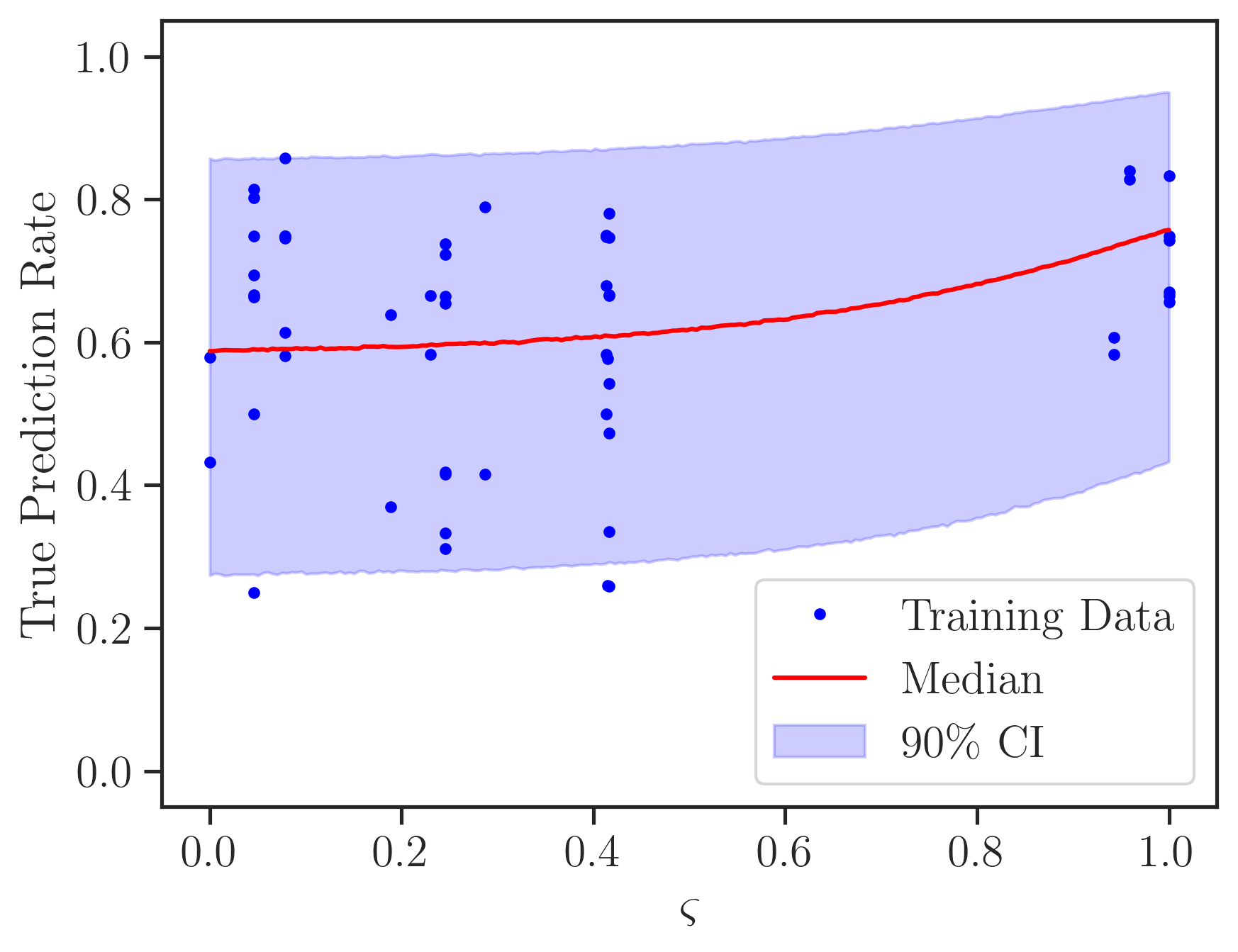}
		\setcounter{subfigure}{0}%
		\caption{}\label{fig:TR}
	\end{subfigure}
	\begin{subfigure}[b]{.45\linewidth}
		\includegraphics[width=\linewidth]{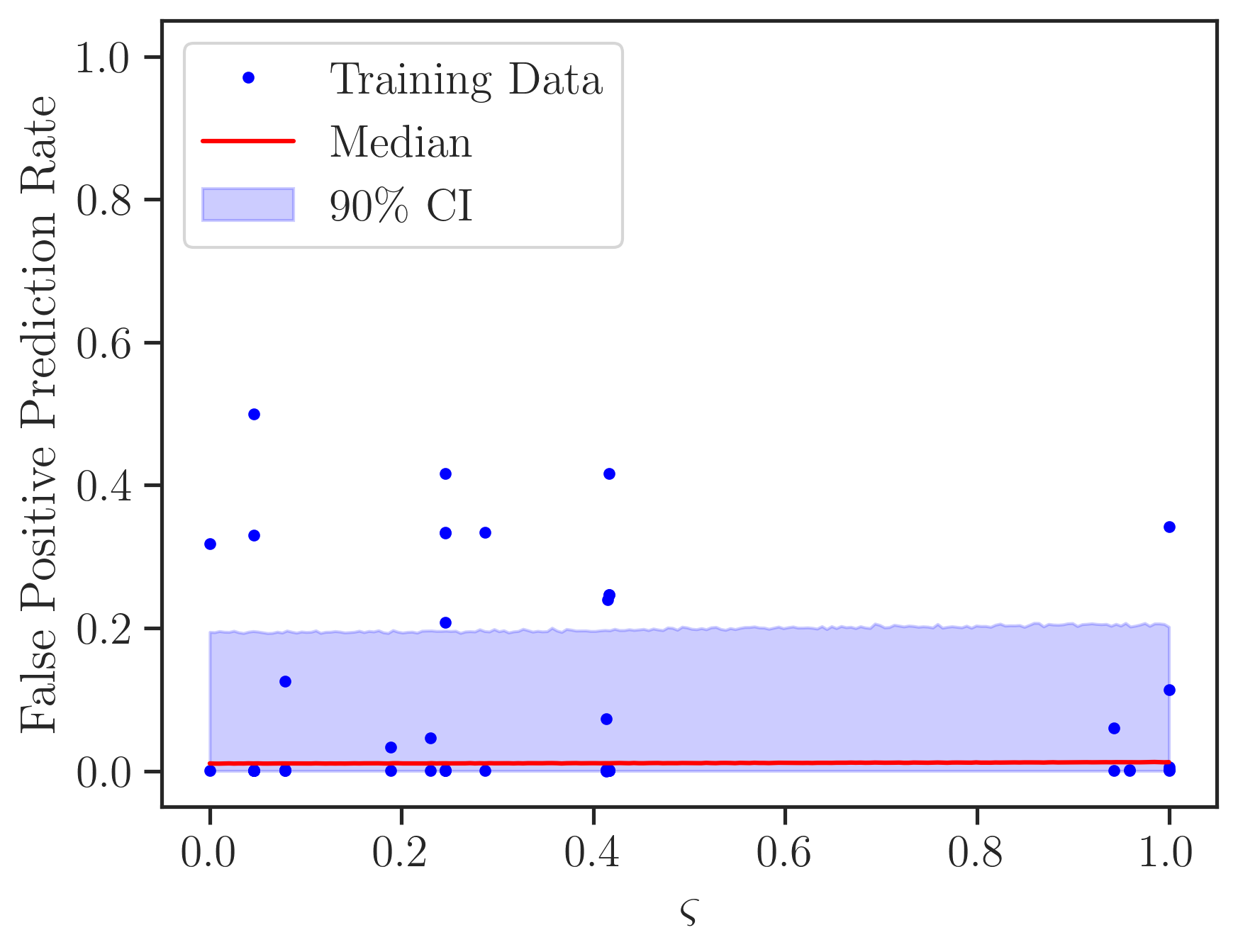}
		\caption{}\label{fig:FPR}
	\end{subfigure}
	
	\begin{subfigure}[b]{.45\linewidth}
		\includegraphics[width=\linewidth]{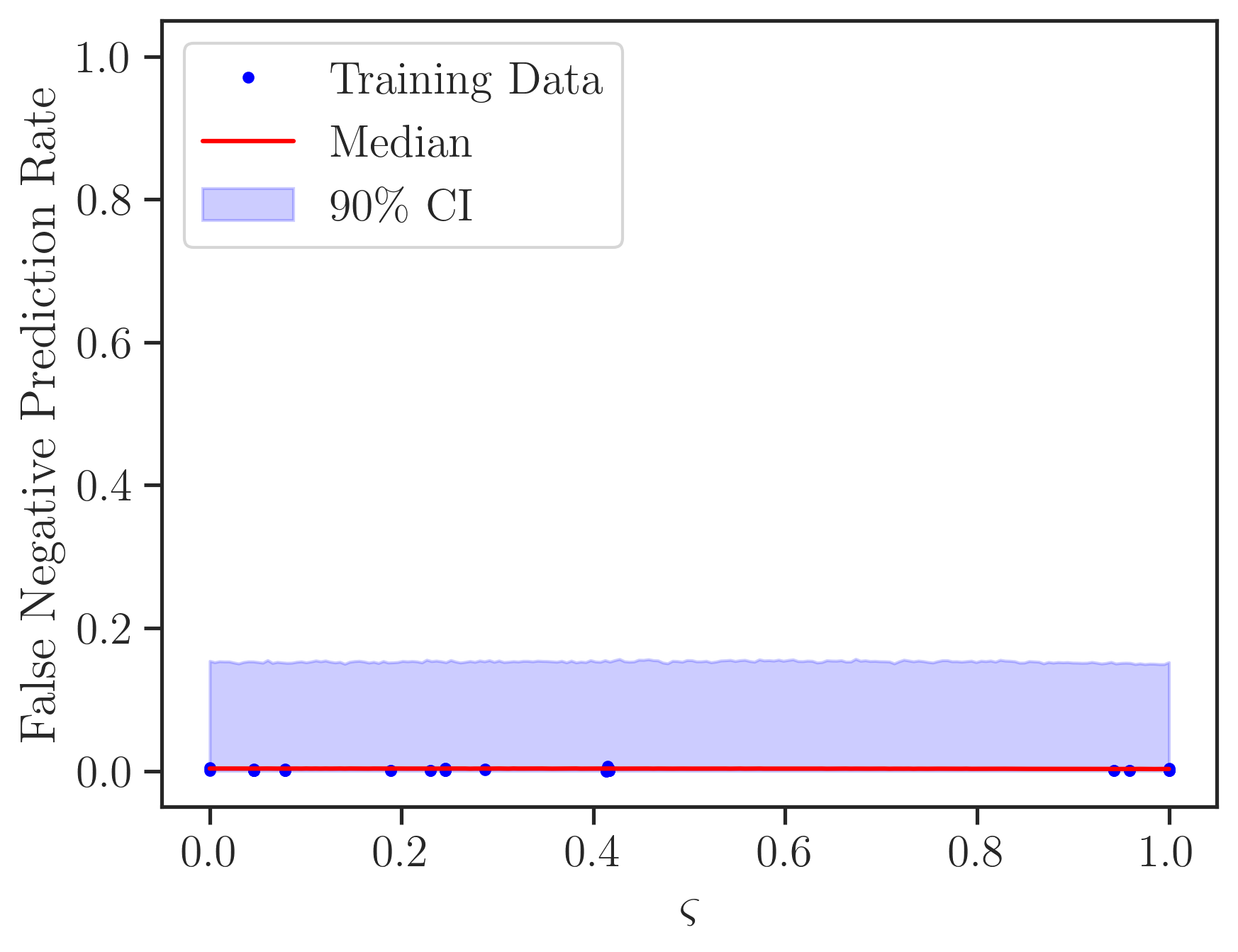}
		\caption{}\label{fig:FNR}
	\end{subfigure}
	\begin{subfigure}[b]{.45\linewidth}
		\includegraphics[width=\linewidth]{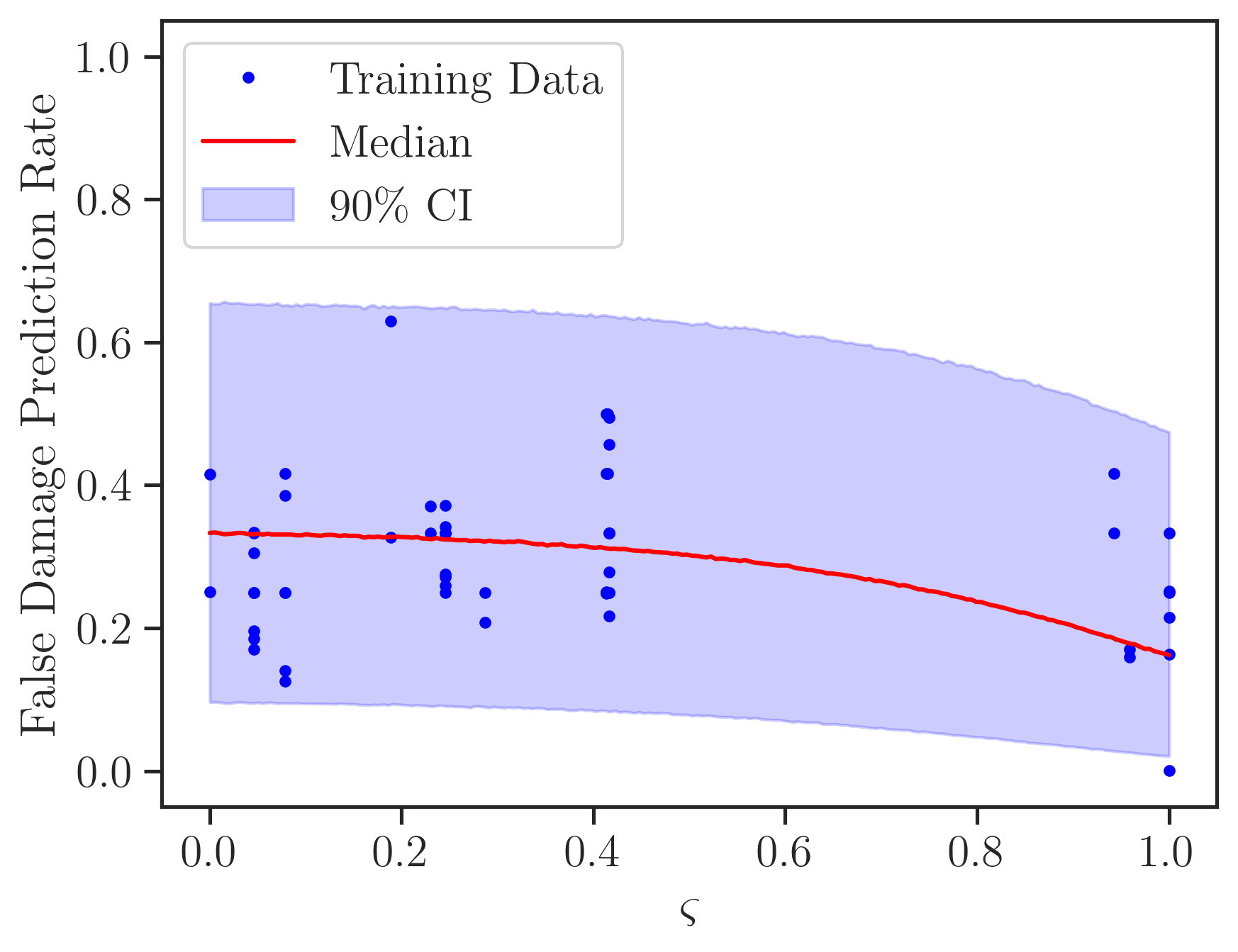}
		\caption{}\label{fig:FDR}
	\end{subfigure}
	\caption{Learned probabilistic functions $p(q_k|\varsigma)$ for (a) the true prediction rate, (b) the false-positive prediction rate, (c) the false-negative prediction rate, and (d) the false-damage prediction rate. Blue shaded regions indicate the 90\% confidence intervals (CI) computed using samples taken from the Dirichlet distributions.}
	\label{fig:results}
\end{figure}

The MLP outlined in the previous section was trained on the data generated from the 56 different transfers possible within the GARTEUR population, using the Adam optimiser \cite{Kingma2014adam}. The results of this probabilistic vector-valued regression are shown in Figure \ref{fig:results}.

From Figure \ref{fig:TR} it can be seen that there is some degree of positive correlation between the TR and the similarity score $\varsigma$. This is of course expected because of the optimisation procedure outlined in the previous section that sought to maximise the Pearson correlation coefficient between TR and $\varsigma$. Fairly high variance in predictive performance is observed for transfers with lower similarity scores; this can be explained by the fact that some structures within the population exhibit low separability between damage classes. Nonetheless, it is apparent from Figure \ref{fig:TR} that the regression successfully learned a monotonically-increasing function mapping from $\varsigma$ to TR which reflects the belief that higher similarity between source and target domain yield more favourable transfer outcomes.

Figures \ref{fig:FPR}, \ref{fig:FNR}, and \ref{fig:FDR} show the misclassification rates versus structural similarity. It can be seen from Figure \ref{fig:FPR} that many transfers results in a false positive prediction rate close to zero, however, there is still some degree of variability. In comparison, there is very little variability in the false negative rate, shown in Figure \ref{fig:FNR}, which is consistently close to zero. The false damage prediction rate, shown in Figure \ref{fig:FDR}, exhibits a weak negative correlation with the similarity score $\varsigma$. Together, these results indicate that transfer results in high rates of damage detection (TR + FDR); though for some cases, localisation remains a challenge. Overall, the probabilistic functions capture the median trends of the data well; however, in some areas the variability is not captured well; specifically, the variability in FPR is underestimated and the variability in FNR is overestimated. This may be a result of lacking enough data points, and could potentially be improved with modifications to the loss function.

\begin{figure}[h!]
	\centering
	\includegraphics[width=0.45\linewidth]{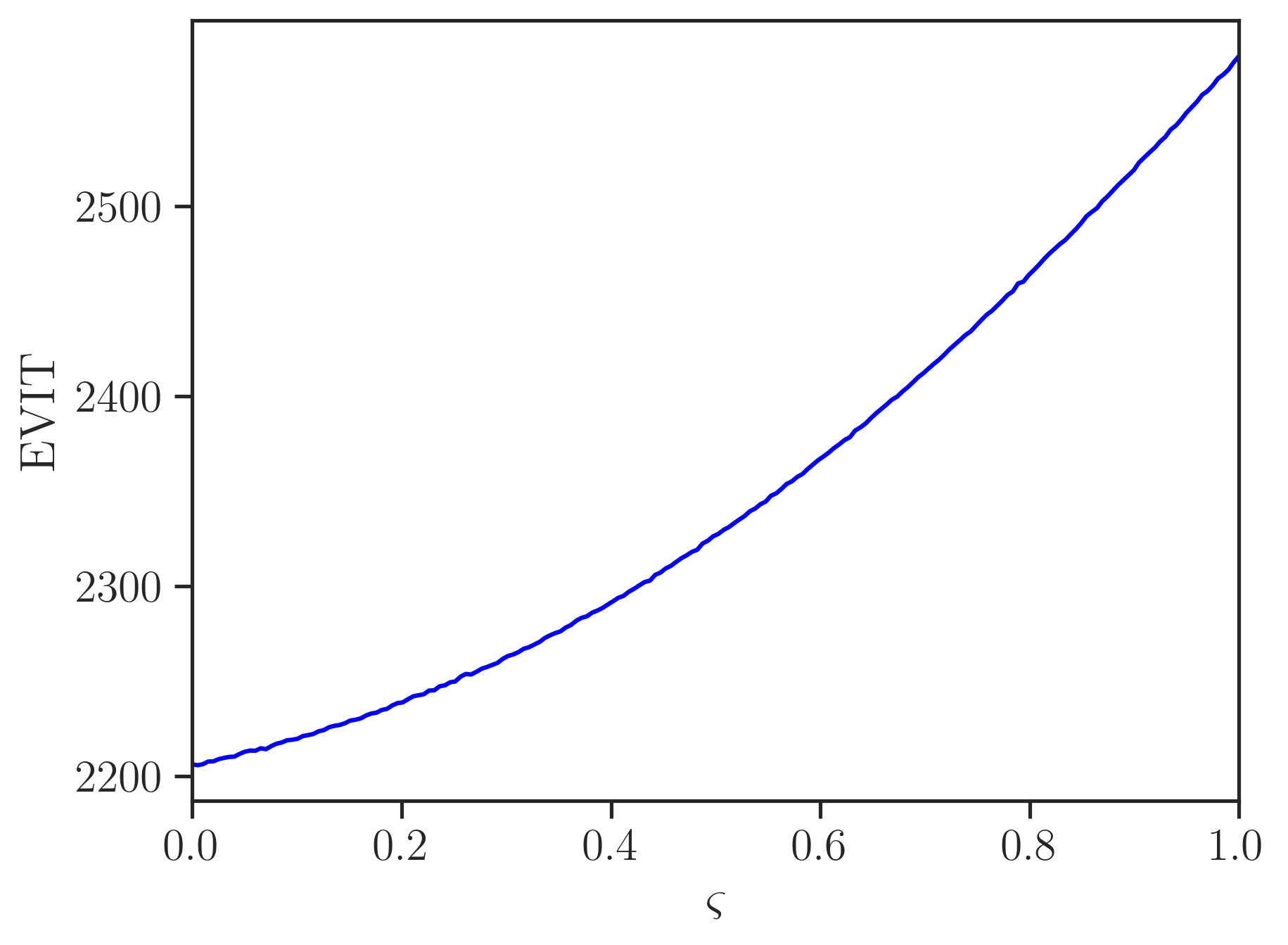}
	\caption{The expected value of information transfer as a function of structural similarity.}
	\label{fig:EVIT}
\end{figure}

Suppose that one acquires a new target domain comprising $M=200$ unlabelled datapoints. By taking the product of samples of the prediction rates from the Dirichlet distribution with $M$, one obtains a distribution over $\bm{q}$. Taking the product of this distribution of the number of each prediction type with the utility function given in Table \ref{tab:UQ} yields a probability distribution over the utility associated with each prediction type. Taking the expectation of this distribution, and summing over the prediction types yields the expected utility $\text{EU}(\mathcal{Q}|\mathcal{T})$ --- the first of the terms required to calculate the EVIT per equation (\ref{eq:EVIT}). The second term, $\text{EU}(\mathcal{Q}|\mathcal{T}_0)$, corresponds to the expected utility associated with the prediction-quality measures following the application of a null transfer strategy, i.e.\ following no transfer. In the current case study, `predictions' were made for the target domain in the null transfer case by guessing labels randomly with equal probability. This assumption yielded $p(\bm{q}|\mathcal{T}_0) = \{0.25, 0.1875, 0.1875, 0.375\}$ and a constant value of $\text{EU}(\mathcal{Q}|\mathcal{T}_0)=-2375$. The EVIT as a function of similarity is shown in Figure \ref{fig:EVIT}.

In Figure \ref{fig:EVIT}, it can be seen that EVIT is a monotonically-increasing function of $\varsigma$, with the maximum EVIT occuring at $\varsigma=1$. This information means that, for any given target domain, the optimal source domain for performing information transfer is the candidate with the highest similarity score when compared to the target domain in question. It can also be seen from Figure \ref{fig:EVIT}, that there are no instances of negative transfer within the population as $\text{EVIT} > 0 \, \forall \, \varsigma$; in essence, this means that transferring from any structure is preferable to no transfer. This result is somewhat expected as the individuals within the GARTEUR population are all quite similar to one another, and it was found in Figure \ref{fig:results} that for many transfer cases, damage detection is improved, if not damage localisation.

To summarise, for the experimental case study of the GARTEUR population, probabilistic functions were learned to map from similarity to transfer efficacy in terms of predictive performance --- this was achieved by assuming a model based on Dirichlet distributions and performing a regression using an MLP. These mappings were then used to predict the performance for potential candidate target domains which further allowed the computation of EVIT. As discussed, this information can be used to optimise transfer learning. 

\section{Conclusions}

Population-based SHM provides a framework for overcoming the limitations of traditional data-based SHM associated with data-scarcity. These benefits are achieved by using information-sharing approaches such as transfer learning and multi-task learning. As decision-support tools, it is important that excellent predictive performance of the statistical models underpinning monitoring systems is maintained lest resources be wasted or structural integrity compromised  --- for this reason, it is paramount for practitioners to maximise the efficacy of transfer learning when applied in the context of PBSHM. The current paper presents an experimental case study based on a population of laboratory-scale aircraft models to demonstrate a process by which the EVIT can be quantified. It was found that for the GARTEUR population, all possible transfers yielded positive value when compared to an uninformed approach to classification in the target domains. Moreover, it was found that the EVIT was a monotonic function of similarity, meaning that for a new target domain, an information transfer strategy can be optimised by selecting the source domain with highest similarity. The quantification of the value provided by information transfer in O\&M decision-making processes provides impetus for continued research into the field of PBSHM.

\section*{Acknowledgements}

The authors would like to gratefully acknowledge the support of the UK Engineering and Physical Sciences Research Council (EPSRC) via grant reference EP/W005816/1. For the purposes of open access, the authors have applied a Creative Commons Attribution (CC BY) license to any Author Accepted Manuscript version arising.

\section*{References}
\vspace{-9mm}
\bibliographystyle{iwshm}
\bibliography{EWSHM2024}

%
%
%
%

\end{document}